\documentclass[10pt,twocolumn,letterpaper]{article}

\usepackage{cvpr}
\usepackage{times}
\usepackage{epsfig}
\usepackage{graphicx}
\usepackage{amsmath}
\usepackage{amssymb}

\usepackage[breaklinks=true,bookmarks=false]{hyperref}

\cvprfinalcopy 

\def\cvprPaperID{5215} 

\begin{document}

\title{DFANet: Deep Feature Aggregation for Real-Time Semantic Segmentation}

\author{Hanchao Li\thanks  {The first two authors contribute equally to this work. This work is done when Hanchao Li is an intern at Megvii Technology.} , Pengfei Xiong\footnotemark[1] , Haoqiang Fan, Jian Sun\\
Megvii Technology\\
{\tt\small \{lihanchao, xiongpengfei, fhq, sunjian\}@megvii.com}
}

\maketitle

\begin{abstract}
   This paper introduces an extremely efficient CNN architecture named DFANet for semantic segmentation under resource constraints. Our proposed network starts from a single lightweight backbone and aggregates discriminative features through sub-network and sub-stage cascade respectively. Based on the multi-scale feature propagation, DFANet substantially reduces the number of parameters, but still obtains sufficient receptive field and enhances the model learning ability, which strikes a balance between the speed and segmentation performance. Experiments on Cityscapes and CamVid datasets demonstrate the superior performance of DFANet with 8$\times$ less FLOPs and 2$\times$ faster than the existing state-of-the-art real-time semantic segmentation methods while providing comparable accuracy. Specifically, it achieves 70.3\% Mean IOU on the Cityscapes test dataset with only 1.7 GFLOPs and a speed of 160 FPS on one NVIDIA Titan X card, and 71.3\% Mean IOU with 3.4 GFLOPs while inferring on a higher resolution image.
\end{abstract}


\section{Introduction}
Semantic segmentation, which aims to assign dense labels for all pixels in the image, is a fundamental task in computer vision. It has a number of potential applications in the fields of autonomous driving, video surveillance, robot sensing and so on. For most such applications, how to keep efficient inference speed and high accuracy with high-resolution images is a critical question.

Previous real-time semantic segmentation approaches \cite{segnet}\cite{SQ}\cite{twocolumn}\cite{BiSeNet}\cite{ICNet}\cite{enet} have already obtained promising performances on various benchmarks\cite{VOC}\cite{CityScapes}\cite{COCO}\cite{ADE20K}\cite{CamVid}. 
However, the operations on the high-resolution feature maps consume significant amount of time in the U-shape structures. 
Some works reduce the computation complexity by restricting the input image size\cite{twocolumn}, or pruning the redundant channels in the network to boost the inference speed\cite{segnet}\cite{enet}. Though these methods seem effective, they easily lose the spatial details around boundaries and small objects. Also, a shallow network weakens feature discriminative ability. 
In order to overcome these drawbacks, other methods \cite{ICNet}\cite{BiSeNet} adopt a multi-branch framework to combine the spatial details and context information. Nevertheless, the additional branches on the high-resolution image limit the speed, and the mutual independence between branches limits the model learning ability in these methods.

\begin{figure}
\begin{center}
   \includegraphics[width=1.0\linewidth]{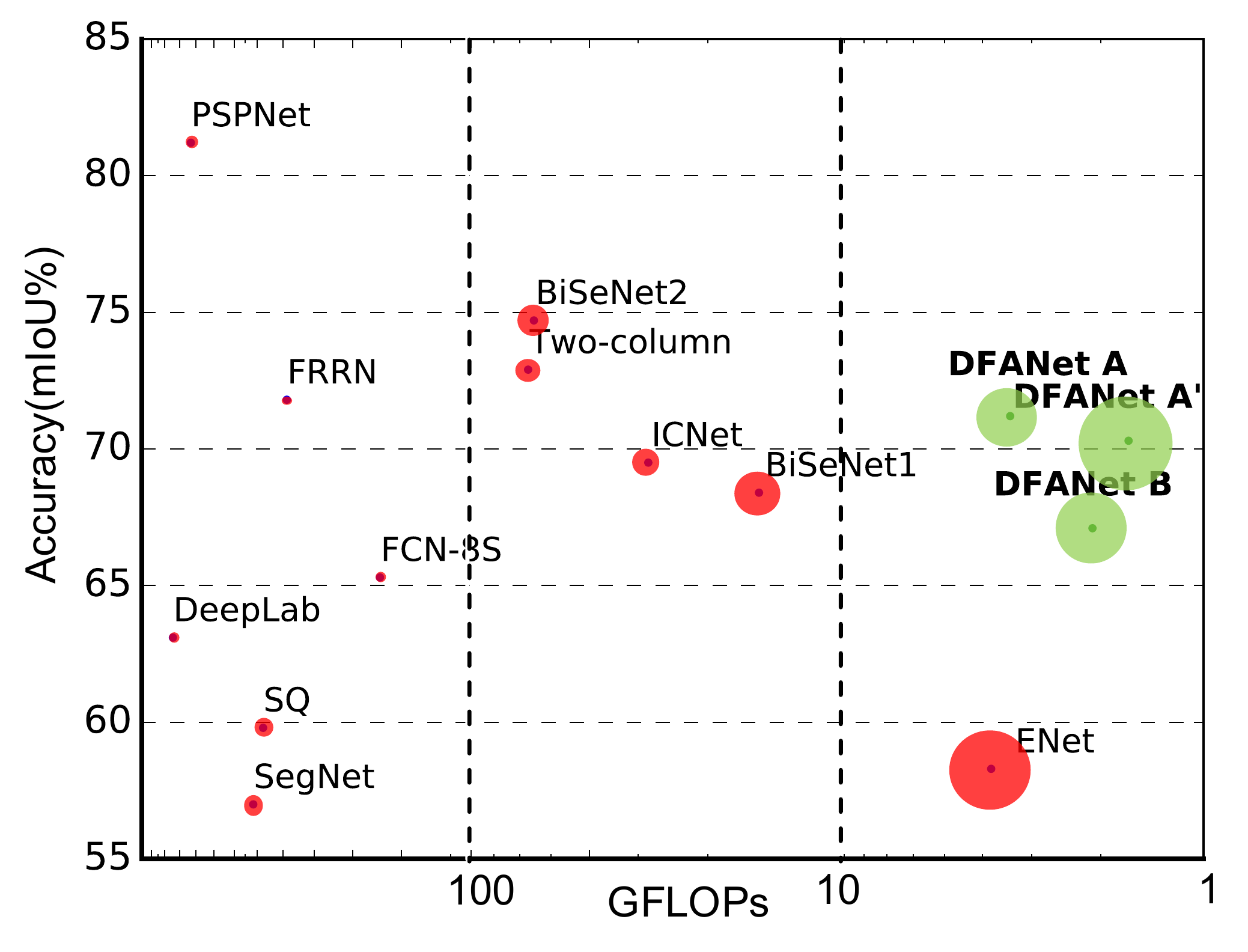}
\end{center}
   \caption{Inference speed, FLOPs and mIoU performance on Cityscapes \textit{test} set. The bigger the circle, the faster the speed.    Results of existing real-time methods, including ICNet\cite{ICNet}, ENet\cite{enet}, SQ\cite{SQ}, SegNet\cite{segnet}, FRRN\cite{frrn}, FCN-8S\cite{fcn_seg}, Two-Column\cite{twocolumn}, BiSeNet\cite{BiSeNet}. Two classical networks DeepLab\cite{deeplabv3plus} and PSPNet\cite{pspnet} are displayed. Also, Our DFANet based on two backbone networks and two input sizes are compared.}
\label{fig:long}
\label{fig:onecol}
\end{figure}


Commonly, semantic segmentation task usually borrows 'funnel' backbone pretrained from image classification task, such as ResNet\cite{ResNet}, Xception\cite{Xception}, DenseNet\cite{DenseNet} and so on. 
For real-time inference, we adopt a lightweight backbone model and investigate how to improve the segmentation performance with limited computation. 
In mainstream semantic segmentation architectures, a pyramid-style feature combination step like Spatial Pyramid Pooling\cite{pspnet}\cite{deeplanv3} is used to enrich features with high-level context, while leading a sharp increase in computational cost.
Moreover, traditional methods usually enrich the feature maps from the final output of a single path architecture. In this kind of design, the high-level context is lacking in incorporation with the former level features which also retain the spatial detail and semantic information in the network path. 
In order to enhance the model learning capacity and increase the receptive field simultaneously, feature reuse is an immediate thought. This motivates us to find a lightweight method to incorporate multi-level context into encoded features.

\begin{figure*}
\begin{center}
\includegraphics[width=1.0\linewidth]{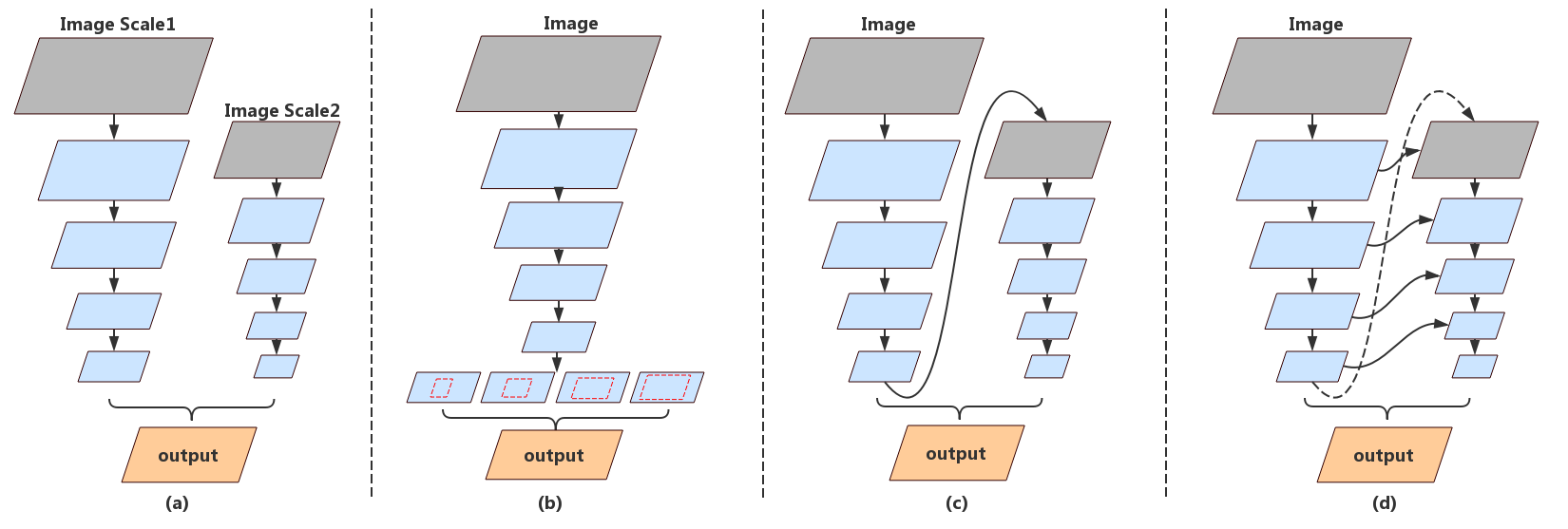}
\end{center}
  \caption{Structure Comparison. From left to right: (a) Multi-branch. (b) Spatial pyramid pooling. (c) Feature reuse in network level. (d) Feature reuse in stage level. As a comparison, the proposed feature reuse methods enrich features with high-level context in another aspect. }
\label{fig:compare_aspp_pan_dfa}
\end{figure*}

In our work, we deploy two strategies to implement cross-level feature aggregation in our model.
First, we reuse high-level features extracted from the backbone to bridge gap between semantic information and structure details.
Second, we combine features of different stages in the processing path of the network architecture to enhance feature representation ability. These ideas are visualized in Figure \ref{fig:compare_aspp_pan_dfa}.

In detail, we replicate the lightweight backbone to verify our feature aggregation methods. Our proposed Deep Feature Aggregation Network (DFANet) contains three parts: the lightweight backbones, sub-network aggregation and sub-stage aggregation modules. Because depthwise separable convolution is proved to be one of the most efficient operation in real-time inference, we modify the Xception network as the backbone structure. In pursuit of better accuracy, we append a fully-connected attention module in the tail of the backbone to reserve the maximum receptive field. 
Sub-network aggregation focuses on upsampling the high-level feature maps of the previous backbone to the input of the next backbone to refine the prediction result. From another perspective, sub-network aggregation can be seen as a coarse-to-fine process for pixel classification. 
Sub-stage aggregation assembles feature representation between corresponding stages through "coarse" part and "fine" part. It delivers the receptive field and high dimension structure details by combining the layers with the same dimension. 
After these three modules, a slight decoder composed of convolution and bilinear upsampling operations is adopted to combine the outputs of each stage to generate the coarse-to-fine segmentation results. The architecture of the proposed network is shown in Figure \ref{fig:whole_net}. 


We test the proposed DFANet on two standard benchmarks, Cityscapes and CamVid. 
With a 1024$\times$1024 input, DFANet achieves 71.3\% Mean IOU with 3.4G FLOPs and speed of 100 FPS on a NVIDIA Titan X card. While implemented on a smaller input size and a lighter backbone, the Mean IOU still stays in 70.3\% and 67.1\% with only 1.7G FLOPs and 2.1G FLOPs respectively, better than most of the state-of-the-art real-time segmentation methods.

Our main contributions are summarized as follows:
\begin{itemize}

\item We set a new record for the real-time and low calculation semantic segmentation. Compared to existing works, our network can be up to 8$\times$ smaller FLOPs and 2$\times$ faster with better accuracy.
\item We present a brand new segmentation network structure with multiple interconnected encoding streams to incorporate high-level context into the encoded features.
\item Our structure provides a better way to maximize the usage of multi-scale receptive fields and refine high-level features several times while computation burden increases slightly. 
\item We modify the Xception backbone by adding a FC attention layer to enhance receptive field with little additional computation.

\end{itemize}

\cvprfinalcopy
\section{Related Work}

\textbf{Real-time Segmentation:} 
Real-time semantic segmentation algorithms are aiming to generate the high-quality prediction under limited calculation. SegNet\cite{segnet} utilizes a small architecture and pooling indices strategy to reduce network parameters. ENet\cite{enet} considers reducing the number of downsampling times in pursuit of an extremely tight framework. Since it drops the last stages of the model, the receptive field of this model is too small to segment larger objects correctly. ESPNet\cite{espnet} performs new spatial pyramid module to make computation efficient. ICNet\cite{ICNet} uses multi-scale images as input and a cascade network to raise efficiency. BiSeNet\cite{BiSeNet} introduces spatial path and semantic path to reduce calculation. Both in ICNet and BiSeNet, only one branch is deep CNN for feature extraction, and other branches are designed to make up resolution details. Different from these methods, we enhance a single model capacity in feature space to reserve more detail information. 

\textbf{Depthwise Separable Convolution:}
Depthwise separable convolution (a depthwise convolution followed by a pointwise convolution), is a powerful operation adopted in many recent neural network designs. This operation reduces the computation cost and the number of parameters while maintaining similar (or slightly better) performance. In particular, our backbone network is based on the Xception model\cite{Xception}, and it shows efficiency in terms of both accuracy and speed for the task of semantic segmentation.

\textbf{High-level Features:} 
The key issues in segmentation task are about the receptive field and the classification ability. In a general encoder-decoder structure, high-level feature of the encoder output depicts the semantic information of the input image. Based on this, PSPNet\cite{pspnet}, DeepLab series\cite{deeplabv3plus}\cite{deeplanv3}\cite{DeepLab2},  PAN\cite{pan} apply an additional operation to combine more context information and multi-scale feature representation. Spatial pyramid pooling has been widely employed to provide a good descriptor for overall scene interpretation, especially for various objects in multiple scales. These models have shown high-quality segmentation results on several benchmarks while usually need huge computing resources.

\textbf{Context Encoding:} 
As SE-Net\cite{senet} explores the channel information to learn a channel-wise attention and has achieved state-of-the-art performance in image classification, attention mechanism becomes a powerful tool for deep neural networks\cite{sca-cnn}. It can be seen as a channel-wise selection to improve module features representation. EncNet\cite{encnet}\cite{mnih2014recurrent}\cite{attentionseg} introduces context encoding to enhance per-pixel prediction that is conditional on the encoded semantics.  In this paper, we also propose a fully-connected module to enhance backbone performance, which has little impact on calculation.

\textbf{Feature Aggregation:} 
Traditional approaches implement a single path encoder-decoder network to solve pixel-to-pixel prediction. As the depth of network increase, how to aggregate features between blocks deserves further attention. 
Instead of simple skip connection design, RefineNet\cite{refinenet} introduces a complicated refine module in each upsampling stage between the encoder and decoder to extract multi-scale features. 
Another aggregation approach is to implement dense connection. The idea of dense connections has been recently proposed for image classification in \cite{DenseNet} and extended to semantic segmentation in \cite{fc-DenseNet} \cite{denseaspp}. DLA\cite{deeplayeraggre} extent this method to develop deeper aggregation structures to enhance feature representation ability.

\section{Deep Feature Aggregation Network}
\begin{figure*}
\begin{center}
\includegraphics[width=1.0\linewidth]{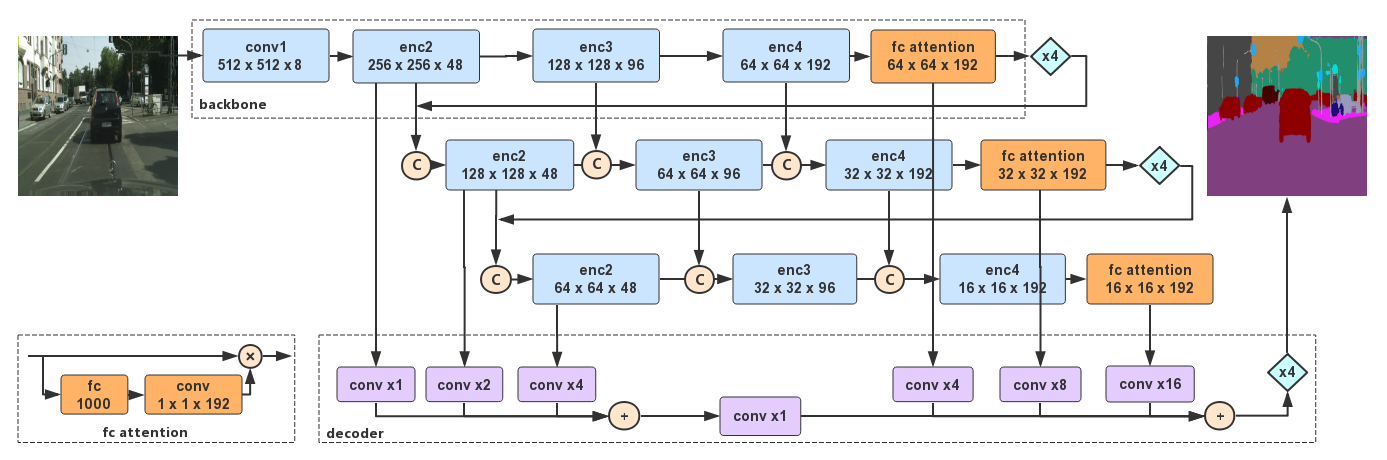}
\end{center}
   \caption{Overview of our Deep Feature Aggregation Network: sub-network aggregation, sub-stage aggregation, and dual-path decoder for multi-level feature fusion. In the figure, "C" means concatenation, "xN" is N$\times$ up-sampling operation.}
\label{fig:whole_net}
\end{figure*}

We start with our observation and analysis of calculation volume when applying current semantic segmentation methods in the real-time task. This motivates our aggregation strategy to combine detail and spatial information in different depth position of the feature extraction network to achieve comparable performance. The whole architecture of \textit{Deep Feature Aggregation Network} (DFANet) is illustrated in Figure \ref{fig:whole_net}.

\subsection{Observations}
We take a brief overview of the segmentation network structures, shown in Figure \ref{fig:compare_aspp_pan_dfa}.

For real-time inference, \cite{ICNet}\cite{BiSeNet} apply multiple branches to perform multi-scale extraction and preserve image spatial details. For example, BiSeNet\cite{BiSeNet} proposed a shallow network process for high-resolution images and a deep network with fast downsampling to strike a balance between classification ability and receptive filed. This structure is displayed in Figure \ref{fig:compare_aspp_pan_dfa}(a).
Nevertheless, the drawback of these methods is obvious that these models are short of dealing with high-level features combined from parallel branches, since it merely implements convolution layers to fuse features. Moreover, features lack communication between parallel branches. Also, the additional branches on high-resolution images limit the acceleration of speed.

In semantic segmentation task, spatial pyramid pooling (SPP) module is a common approach to deal with high-level features
\cite{deeplanv3} (Figure \ref{fig:compare_aspp_pan_dfa}(b)). The ability of spatial pyramid module is to extract high-level semantic context and increase receptive field, such as \cite{DeepLab2}\cite{pspnet}\cite{pan}. However, 
implementing spatial pyramid module is usually time-consuming. 

Inspired by the above methods, we firstly replace the high-level operation 
by upsampling the output of a network and refining the feature map with another sub-network, as shown in Figure \ref{fig:compare_aspp_pan_dfa}(c). Different from SPP module, the feature maps are refined on a larger resolution and sub-pixel details are learned simultaneously. 
However, as the whole structure depth grows, high-dimension features and receptive field usually suffer precision loss since the feature flow is a single path. 

Pushing a bit further, we propose stage-level method (Figure \ref{fig:compare_aspp_pan_dfa}(d)) to deliver low-level features and spatial information to semantic understanding. Since all these sub-networks have the similar structure, stage-level refinement can be produced by concatenating the layers with the same resolution to generate the multi-stage context. Our proposed \textit{Deep Feature Aggregation Network} aims to exploit features combined from both network-level and stage-level.

\subsection{Deep Feature Aggregation}

We focus on making the fusion of different depth features in networks. Our aggregation strategy is composed of sub-network aggregate and sub-stage aggregate methods. The structure of DFANet is illustrated in Figure \ref{fig:whole_net}.

\textbf{Sub-network Aggregation.} Sub-network aggregation implements combination of high-level features at the network level.
Based on the above analysis, we implement our architecture as a stack of backbones by feeding the output of the previous backbone to the next. From another perspective, sub-network aggregation could be seen as a refinement process. A backbone process is defined as $ y = \Phi(x)$, the output of encoder $\Phi_{n}$ is the input of encoder $\Phi_{n+1}$, so sub-network aggregate can be formulated as: $Y = \Phi_n(\Phi_{n -1}(...\Phi_{1}(X)))$.

A similar idea has been introduced in \cite{stack_hourglass}. The structure is composed with a stack of encoder-decoder "hourglass" network. Sub-network aggregation allows these high-level features to be processed again to further evaluate and reassess higher order spatial relationships.

\textbf{Sub-stage Aggregation.} Sub-stage aggregation focuses on fusing semantic and spatial information in stage-level between multiple networks. As the depth of network grows, spatial details suffer losing. Common approaches, like U-shape, implement skip connection to recover image details in the decoder module. However, the deeper  encoder blocks lack low-level features and spatial information to make judgments in large-scale various objects and precise structure edge. Parallel-branch design uses original and decreased resolution as input, and the output is the fusion of large-scale branch and small-scale branch results, while this kind of design has a lack of information communication between parallel branches.

Our sub-stage aggregation is proposed to combine features through encoding period. We make the fusion of different stages in the same depth of sub-networks.
In detail, the output of a certain stage in the previous sub-network is contributed to the input of the next sub-network in the corresponding stage position.

For a single backbone $\Phi_n(x)$, a stage process can be defined as $\phi_{n}^{i}$. The stage in the previous backbone network is $\phi_{n-1}^{i}$. \text{ }$i$ means the index of the stage. Sub-stage aggregation method can be formulated as:




\begin{equation}
x_{n}^i=\left\{\begin{array}{ll}

x_{n}^{i-1} + \phi_{n}^{i}(x_{n}^{i-1})& {\text{if } n = 1},\\ 
{[x_{n}^{i-1},x_{n-1}^{i}] + \phi_{n}^{i}([x_{n}^{i-1},x_{n-1}^{i}])}& {\text{otherwise,}}

\end{array}\right.
\label{equ:sub-stage}
\end{equation}

While, $x_{n-1}^i$ is coming from:
\begin{equation}
x_{n-1}^{i} = x_{n-1}^{i-1} + \phi_{n - 1}^{i}(x_{n-1}^{i-1})
\end{equation}

Traditional approaches are learning a mapping of $\mathcal{F}(x) + x$ for $x_{n}^{i-1}$. In our proposed method, sub-stage aggregation method is learning a residual formulation of ${[x_{n}^{i-1},x_{n-1}^{i}]}$, at the beginning of each stage.

For $n > 1$ situation, the input of $i$th stage in $n$th network is given by combining the $i$th stage output in ($n-1$)th network with the ($i - 1$)th stage output in $n$th network, then the $i$th stage learns a residual representation of $[x_{n}^{i-1},x_{n-1}^{i}]$. $x_{n}^{i-1}$ has the same resolution as $x_{n-1}^{i}$, and we implement concatenation operation to fuse features.

We keep the feature always flow from high-resolution into the low-resolution.
Our formulation not only learns a new mapping of $n$th feature maps but also preserves ($n - 1$)th features and receptive field. Information flow can be transferred through multiple networks.

\subsection{Network Architecture}
\label{sec:fc}

The whole architecture is shown in Figure \ref{fig:whole_net}. In general, our semantic segmentation network could be seen as an encoder-decoder structure. As discussed above, the encoder is an aggregation of three Xception backbones, composed with sub-network aggregate and sub-stage aggregate methods. For real-time inference, we don't put too much focus on the decoder. The decoder is designed as an efficient feature upsampling module to fuse low-level and high-level features.
For convenience to implement our aggregate strategy, our sub-network is implemented by a backbone with single bilinear upsampling as a naive decoder.
All these backbones have the same structure and are initalized with same pretrained weight.

\textbf{Backbone.} The basic backbone is a lightweight Xception model with little modification for segmentation task, we will discuss the network configuration in the next section. For semantic segmentation, not only providing dense feature representation, how to gain semantic context effectively remains a problem. Therefore, we preserve fully-connected layers from ImageNet pretraining to enhance semantic extraction. 
In classification task, fully-connected (FC) layer is followed by global pooling layers to make final probability vectors. Since classification task dataset \cite{imagenet} provides large amount of categories than segmentation datasets \cite{VOC}\cite{ADE20K}. Fully-connected layer from ImageNet pretraining could be more powerful to extract category information than training from segmentation datasets. 
We apply a $1\times1$ convolution layer followed with FC layer to reduce channels to match the feature maps from Xception backbone. 
Then $N\times C\times1\times1$ encoding vector is multiplied with original extracted features in channel-wise manner.

\textbf{Decoder.} Our proposed decoder module is illustrated in Figure \ref{fig:whole_net}. For real-time inference, we don't put too much focus on designing complicated decoder module. According to DeepLabV3+\cite{deeplabv3plus}, not all the features of the stages are necessary to contribute to decoder module. We propose to fuse high-level and low-level features directly. Because our encoder is composed of three backbones, we firstly fuse high-level representation from the bottom of three backbones. Then the high-level features are bilinearly upsampled by a factor of 4, and low-level information from each backbone that have the same spatial resolution is fused respectively. Then the high-level features and low-level details are added together and upsampled by a factor of 4 to make the final prediction. In decoder module, we only implement a few convolution calculations to reduce the number of channels.

\section{Experiments}
While our proposed network is effective for high resolution images, we evaluate it on two challenging benchmarks: \textbf{Cityscapes} and \textbf{CamVid}. The image resolution of these two datasets are up to $2048\times1024$ and $960\times720$ respecitivaly, which makes it a big challenge for real-time semantic segmentation. In the following, we first investigate the effects of the proposed architecture, then conduct the accuracy and speed results on Cityscapes and CamVid compared with the existing real-time segmentation algorithms. 

All the networks mentioned below follow the same training strategy. They are trained using mini-batch stochastic gradient descent (SGD) with batch size 48, momentum 0.9 and weight decay $1e-5$. As common configuration, the "poly" learning rate policy is adopted where the initial rate is multiplied by $(1-\frac{iter}{max\_iter})^{power}$ with power 0.9 and the base learning rate is set as $2e-1$. The cross-entropy error at each pixel over the categories is applied as our loss function. Data augmentation contains mean subtraction, random horizontal flip, random resizing with scale ranges in [0.75, 1.75], and random cropping into fix size for training.

\subsection{Analysis of DFA Architecture}

We adopt Cityscapes to conduct the quantitative and qualitative analysis of experiments firstly. The Cityscapes is comprised of a large, diverse set of stereo video sequences recorded in streets from 50 different cities, containing 30 classes, and 19 of them are considered for training and evaluation. The dataset contains 5,000 finely annotated images and 19,998 images with coarse annotation, which all have a high resolution of $2048\times1024$. Following the standard setting of Cityscapes, the fine annotated images are split into training, validation and testing sets with 2,979, 500 and 1,525 images respectively. We only use the fine annotated images during training and stop the training process after 40K iterations. 

\begin{table}
\begin{center}
\begin{tabular}{l|c|c}
\hline
stage & Xception A & Xception B   \\
\hline\hline
\text{conv1} & $3\times3$, 8, stride 2 & $3\times3$, 8, stride 2   \\
\hline
enc2              & $\left[\begin{array}{c}
\text{ $3\times3$, } 12 \\
\text{ $3\times3$, } 12 \\
\text{ $3\times3$, } 48
\end{array} \right] \times 4$ & $\left[\begin{array}{c}
\text{ $3\times3$, } 8 \\
\text{ $3\times3$, } 8 \\
\text{ $3\times3$, } 32
\end{array} \right] \times 4$  \\
\hline
enc3              & $\left[\begin{array}{c}
\text{ $3\times3$, } 24 \\
\text{ $3\times3$, } 24 \\
\text{ $3\times3$, } 96
\end{array} \right] \times 6$ & $\left[\begin{array}{c}
\text{ $3\times3$, } 16 \\
\text{ $3\times3$, } 16 \\
\text{ $3\times3$, } 64
\end{array} \right] \times 6$  \\
\hline
enc4              & $\left[\begin{array}{c}
\text{ $3\times3$, } 48 \\
\text{ $3\times3$, } 48\\
\text{ $3\times3$, } 192
\end{array} \right] \times 4$ & $\left[\begin{array}{c}
\text{ $3\times3$, } 32 \\
\text{ $3\times3$, } 32 \\
\text{ $3\times3$, } 128
\end{array} \right] \times 4$  \\
\hline

\hline
\end{tabular}
\end{center}
\caption{Modified Xception architecture. Building blocks are shown in brackets with the numbers of blocks stacked. $3\times3$ means a \textit{depthwise separable convolution} except "conv1". In "conv1" stage, we only implement a $3\times3$ convolution layer.
}
\label{tab:backbone-struct}
\end{table}

The model performance is evaluated on Cityscapes validation set. For fair comparison, we make the ablation study under $1024\times1024$ crop size. In this process, we don't employ any testing augmentation, like multi-scale or multi-crop testing for the best result quality. For quantitative evaluation, the mean of class-wise intersection over union (mIoU), and the number of float-point operations (FLOPs) are applied to investigate the accuracy and computation complexity measurement respectively. 

\subsubsection{Lightweight Backbone Networks}
As mentioned above, backbone network is one of the major limitations of model acceleration. However, too small backbone networks lead to serious degradation of segmentation accuracy. Xception, designed with lightweight architecture, is known as achieving better speed-accuracy tradeoff. We implement two modified Xception network (Xception A, Xception B) with even less computation complexity to pursue the inference speed  of our proposed method. The detailed architectures of these two models are summarized in Table \ref{tab:backbone-struct}. 

The proposed Xception networks are pretrained on ImageNet-1k dataset with similar training protocol in \cite{imagenet}\cite{deeplabv3plus}. Specifically, we adopt Nesterov momentum optimizer with momentum = 0.9, initial learning rate = 0.3, and weight decay $4e-5$. After training with 30 epoches, we set learning rate $= 0.03$ for another 30 epoches. Our batch size is 256 and image size is $224 \times 224$. We did not tune the hyper-parameters very hard as the goal is to pretrain the model on ImageNet for semantic segmentation. 

We evaluate proposed modified Xception on Cityscapes val dataset. To make prediction resolution equal with original images, the features are bilinearly upsampled by a factor of 16. Taken as comparison, we reproduce ResNet-50, which adopts dilated convolution to make 1/16 downsample. As can be seen, when taking Xception A instead of ResNet-50, the segmentation accuracy decrease from 68.3\% to 59.2\%.  However, the performance decreases less when implementing with ASPP\cite{deeplanv3} (72.1\% of ResNet-50 + ASPP $\rightarrow$ 67.1\% of Xception A + ASPP), which proves the effectiveness of ASPP module on lightweight backbone. Followed by ASPP module, Xception A achieves 67.1\% mIoU, which is comparable with 68.3\% of ResNet-50, while the computational complexity of the former is far less than the latter. That supports us to apply a lightweight model accompanied by a high-level contextual module for semantic segmentation under resource constraints. 

We also consider decreasing the resolution of input images to accelerate computation. In the previous methods, researchers try to apply the lower resolution input to achieve real-time inference. However, when scaling ratio is 0.25, the corresponding mIoU is intolerably low. While inferring with a much smaller size input, the FLOPs of original model is still markedly bigger than a small backbone (9.3G of ResNet-50 $\rightarrow$ 1.6G of Xception A). With the ASPP following, Xception A easily achieves better accuracy than the traditional ResNet-50. Even applied on another smaller backbone Xception B, the accuracy is comparable and the FLOPs is half. Despite the usefulness of ASPP module, the computational complexity is obviously too large. As an alternative to the global pooling attention module, we evaluate the influence of FC attention module introduced in Section \ref{sec:fc}. As shown in Table \ref{tab: backbone comparison}, for both Xception A and B, FC attention can gain $4-6\%$ accuracy improvement, which is notable while the amount of computation is almost unchanged. FC attention provides evidence for the effect of high-dimensional context, and implements a simple and effective method to fuse the image contextual information from a global perspective. In the following experiments, we take Backbone A and B as our basic unit to construct the performance of our DFANet.

\begin{table}
\begin{center}
\begin{tabular}{l|c|c|c}
\hline
Model & Scale & FLOPs & mIoU(\%) \\
\hline\hline
ResNet-50 & 0.25 & 9.3G              & 64.5 \\
ResNet-50 & 1.0 &  149.2G             & 68.3 \\
\hline
ResNet-50 + ASPP & 1.0 & 214.4G      & 72.1 \\
\hline\hline

\hline
Xception A          & 1.0    & 1.6G         &  59.2 \\
Xception A + ASPP   & 1.0    & 6.9G         &  67.1 \\
Xception B          & 1.0    & 0.83G        &  55.4 \\
Xception B + ASPP   & 1.0    & 4.4G         &  64.7 \\ 
\hline\hline
Backbone A  & 1.0 & 1.6G    &   65.4  \\
Backbone B  & 1.0 & 0.83G   &   59.2  \\ 
\hline
\end{tabular}
\end{center}
\caption{Different structure followed with or without ASPP, evaluate on Cityscapes \textit{val} dataset. 
'Backbone' means Xception network followed with FC attention module. 'Scale' means scaling ratio of input image.}
\label{tab: backbone comparison}
\end{table}

\subsubsection{Feature Aggregation}
In this subsection, we investigate the effect of aggregation strategy in our proposed network. Our feature aggregation is composed of sub-network aggregation and sub-stage aggregation. We replicate backbones to show the performance on Cityscapes \textit{val} set.

As shown in Table \ref{tab: feature aggregate comparison}, based on the proposed Backbone A, the segmentation accuracy is improved from 65.4\% to 66.3\%, while applying sub-network aggregation once. When applying aggregation twice('$\times3$'), the accuracy is slightly decreased from 66.3\% to 65.1\%. We think that the receptive field of Backbone A x2 is already bigger than the whole image, so another aggregation introduces some noise. As the output is directly upsampled to the original size, the noise is amplified as well. Although it brings more details, noise also brings negative interference. When aggregation number is '$\times4$', we don't gain much benefit on the accuracy. Because the final output resolution is $8\times8$ when input resolution is $1024\times1024$, the features are too small to make category classification. 

\begin{table}
\begin{center}
\begin{tabular}{l|c|c|c}
\hline
Model & FLOPs & Params & mIoU(\%) \\
\hline\hline
Backbone A     & 1.6G & 2.1M & 65.4 \\
Backbone A x2  & 2.4G & 4.9M & 66.3 \\
Backbone A x3  & 2.6G & 7.6M & 65.1 \\
Backbone A x4  &2.7G & 10.2M & 50.8 \\
\hline
Backbone B     & 0.83G & 1.4M & 59.2 \\
Backbone B x2  & 1.2G & 3.1M & 62.1 \\
Backbone B x3  & 1.4G & 4.7M & 58.2 \\
Backbone B x4  & 1.5G & 6.3M & 50.7 \\
\hline
\end{tabular}
\end{center}
\caption{Detailed performance comparison of our proposed aggregation strategy. '$\times N$' means that we replicate N backbones to implement feature aggregation.}
\label{tab: feature aggregate comparison}
\end{table}

\begin{figure}
\begin{center}
   \includegraphics[width=1.0\linewidth]{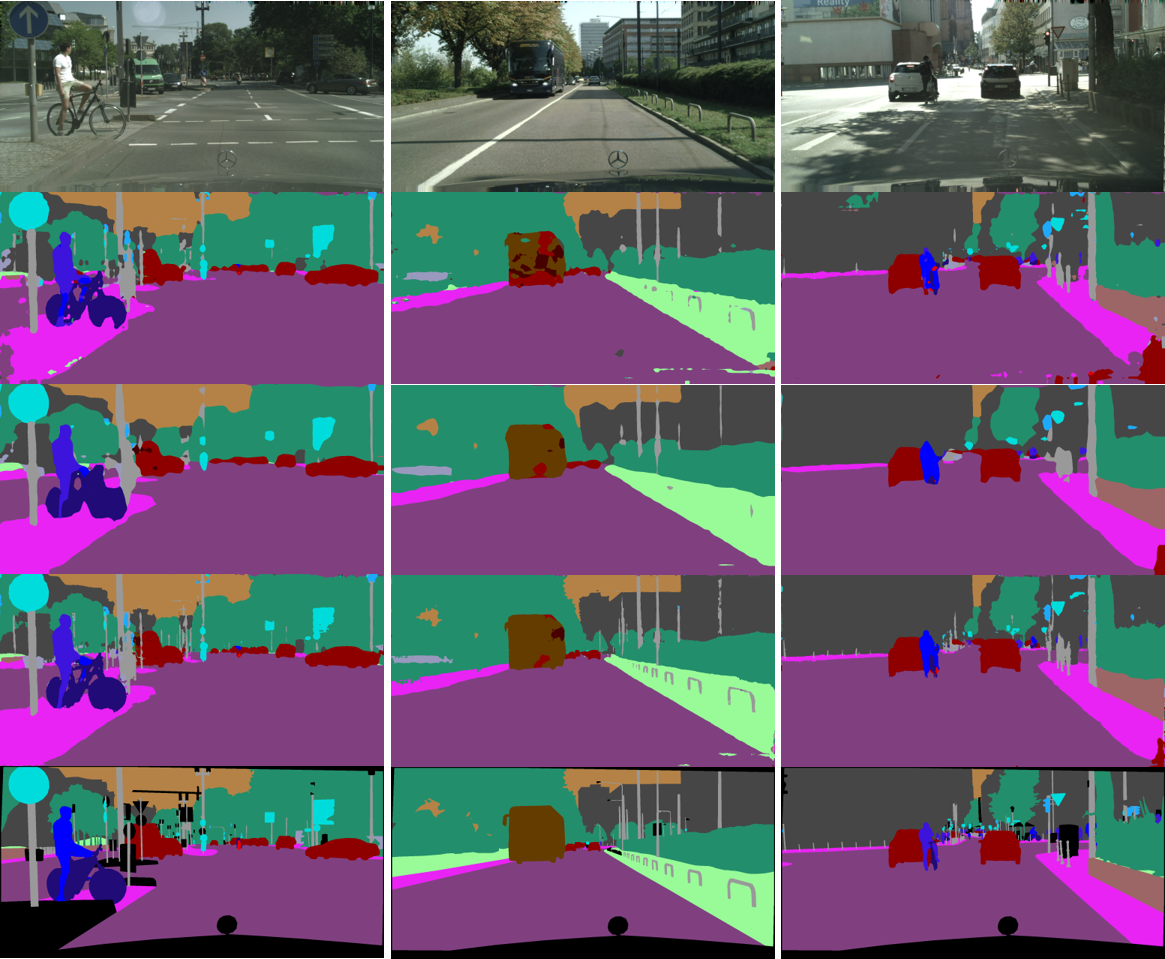}
\end{center}
   \caption{Results of the proposed DFANet on Cityscapes validation set. The first line is input images, and Line 2$\sim$4 display the output of each backbone in DFANet. The final line is ground truth.}
\label{fig:visual results}
\end{figure}


\begin{table}
\begin{center}
\begin{tabular}{l|c|c|c}
\hline
Model & FLOPs & Params & mIoU \\
\hline\hline
Backbone A x2        & 2.4G & 4.9M & 66.3 \\
Backbone A x2+HL     & 2.5G & 5.0M & 67.1 \\
Backbone A x2+HL+LL  & 3.2G & 5.1M & 69.4 \\
\hline
Backbone A x3        & 2.6G & 7.6M & 65.1 \\
Backbone A x3+HL     & 2.7G & 7.7M & 69.6 \\
Backbone A x3+HL+LL  & 3.4G & 7.8M & 71.9 \\
\hline
Backbone B x3        & 1.4G & 4.7M & 58.2 \\
Backbone B x3+HL     & 1.5G & 4.9M & 67.6 \\
Backbone B x3+HL+LL  & 2.1G & 4.9M & 68.4 \\
\hline
\end{tabular}
\end{center}
\caption{Detailed performance comparison of our proposed decoder module. 'HL' means that fusing high-level features. 'LL' means fusing low-level features.}
\label{tab:decoder results}
\end{table}

Figure \ref{fig:visual results} displays the results of three stacked backbones. As can be seen, the prediction of first backbone has a lot of noise, then it becomes smoother in the next stage with spatial detail corruption. This result proves that the receptive field is enlarged and global context is introduced after sub-stage learning. Then, processed by the third aggregation backbone, the structure details become more precise in the result. Both the detail and contextual information are combined in the prediction result after the third refinement. We believe that, sub-stage aggregation brings the combination of multi-scale information. Based on our cascaded model, more discriminative features are learned, and sub-pixel learning is processed progressively. 


\subsubsection{The Whole DFA Architecture}
Finally, we conduct the whole results of the proposed DFA architecture. In Section \ref{sec:fc}, our decoder module is designed as effective and simple to combine high-level and low-level features. Different from directly upsampling, the convolutions in decoder module further smooth the combined results. The performance of the aggregation encoder is shown in Table \ref{tab:decoder results}. 

Although the performance of Backbone A x3 is slightly worse than Backbone A x2, the final aggregation encoder is composed of three backbones, as shown in Figure \ref{fig:whole_net}. Based on the decoder operation, the accuracy of Backbone A x3 is much better than Backbone A x2. As with the previous conclusion, it also illustrates that details are learned in sub-stage 3, while noises are ablated in the combination of different scale outputs. 

Since our aggregation methods can provide dense features, we do not pursue complicated decoder module design as inference speed requirements. Based on the two types of backbones, all of the high-level and low-level decoders have further improved the performance with slight increase in computational effort. 
Based on all the above analysis, we obtain the final result on Cityscapes \textit{val} set with 71.9\% mIoU and only 3.4 GFLOPs. Furthermore, the computation of the whole architecture based on Backbone B is decreased to 2.1 GFLOPs, but it still achieves 68.4\% mIoU.

\subsection{Speed and Accuracy Comparisons}

\begin{table*}
\begin{center}
\begin{tabular}{|l|c|c|c|c|c|c|}
\hline
Model                     & InputSize & FLOPs                  & Params & Time(ms) & Frame(fps) & mIoU(\%) \\
\hline\hline
PSPNet\cite{pspnet}      & 713 $\times$ 713    & 412.2G     & 250.8M     & 1288 & 0.78 & 81.2 \\
DeepLab\cite{DeepLab2}    & 512 $\times$ 1024  & 457.8G  & 262.1M     & 4000 & 0.25 & 63.1 \\
\hline
SegNet\cite{segnet}       & 640 $\times$ 360   & 286G    & 29.5M & 16   & 16.7 & 57   \\
ENet\cite{enet}           & 640 $\times$ 360   & 3.8G    & 0.4M  & \textbf{7}    & \textbf{135.4} & 57   \\
SQ\cite{SQ}               & 1024 $\times$ 2048 & 270G    & -     & 60   & 16.7 & 59.8 \\
CRF-RNN\cite{crfasRnn}    & 512 $\times$ 1024  & -       & -     & 700  & 1.4  & 62.5 \\
FCN-8S\cite{fcn_seg}      & 512 $\times$ 1024  & 136.2G  & -     & 500  & 2 & 63.1 \\
FRRN\cite{frrn}           & 512 $\times$ 1024  & 235G      & -     & 469  & 0.25 & 71.8 \\
ICNet\cite{ICNet}         & 1024 $\times$ 2048 & 28.3G   & 26.5M     & 33   & 30.3 & 69.5 \\
TwoColumn\cite{twocolumn} & 512 $\times$ 1024  & 57.2G   & -     & 68   & 14.7 & 72.9 \\
BiSeNet1\cite{BiSeNet}    & 768 $\times$ 1536  & 14.8G   & 5.8M  & 13   & 72.3 & 68.4 \\
BiSeNet2\cite{BiSeNet}    & 768 $\times$ 1536  & 55.3G   & 49M   & 21   & 45.7 & \textbf{74.7} \\
\hline
DFANet A                  & 1024 $\times$ 1024 & \textbf{3.4G}  & 7.8M & \textbf{10} & \textbf{100}  & 71.3 \\
DFANet B                  & 1024 $\times$ 1024 & \textbf{2.1G}  & 4.8M & \textbf{8} & \textbf{120}  & 67.1 \\
DFANet A'                 & 512 $\times$ 1024 & \textbf{1.7G}  & 7.8M & \textbf{6} & \textbf{160}  & 70.3 \\
\hline
\end{tabular}
\end{center}
\caption{Speed analysis on Cityscapes \textit{test} dataset. "-" indicates that the corresponding result is not provided by the methods.}
\label{tab:speed analysis}
\end{table*}

The overall speed comparison is demonstrated in Table~\ref{tab:speed analysis}. Speed is a vital factor of an algorithm, we try to test our model under the same status thorough comparison. The network inference time is applied here to investigate the effectiveness. All experiments are developed on a virtual machine with a single Titan X GPU card. For the proposed method, we report the average time from running through the all test images from Cityscapes using our best performing networks. The resolutions of the input image are also listed for comparison in the table. In this process, we don't employ any testing augmentation.

As can be observed, while the inference speed of the proposed method significantly outperforms state-of-the-art methods, the accuracy performance is kept comparable, attributing to the simple and efficient pipeline. The baseline of the proposed method achieves mIoU 71.3\% on Cityscapes test set with 100 FPS inference speed. We extend the proposed method in two aspects that the input size and the channel dimension. When the backbone model is decreases to a simplied one, the accuracy performance of DFANet is decreased to 67.1\% corresponding with still 120 FPS inference speed, which is comparable with the previous state-of-the-art with 68.4\% of bisenet\cite{BiSeNet}. However, while the height of input image is downsampled to half, the FLOPs of the DFANet A drops to 1.7G, but the accuracy is still good enough to better than several existing methods. The fastest setting of our method runs at a speed of 160 FPS at mIoU 70.3\%, while the previous fastest results\cite{enet} is only 135 FPS at mIoU 57\%. Compared with the previous state-of-the-art model\cite{BiSeNet}, the proposed DFANet A, B, A' has 1.38 $\times$, 1.65 $\times$ and 2.21 $\times$ speed acceleration and only 1/4, 1/7 and 1/8 FLOPs, with even slightly better segmentation accuracy. Some visual results of the proposed DFANet A is showed in Figure \ref{fig:visual results}. With the proposed feature aggregation structure, we produce decent prediction results on Cityscapes. 

\begin{table}
\begin{center}
\begin{tabular}{|l|c|c|c|}
\hline
Model & Time(ms) & Frame(fps) & mIoU(\%) \\
\hline\hline
SegNet\cite{segnet}    & 217 & 46   & 46.4 \\
DPN\cite{dfn}          & 830 & 1.2  & 60.1 \\
DeepLab\cite{DeepLab2} & 203 & 4.9  & 61.6 \\
ENet\cite{enet}        & -   & -    & 51.3 \\
ICNet\cite{ICNet}      & 36  & \textbf{27.8} & 67.1 \\
BiSeNet1\cite{BiSeNet} & -   & -    & 65.6 \\
BiSeNet2\cite{BiSeNet} & -   & -    & \textbf{68.7} \\
\hline
DFANet A               & 8  & \textbf{120}  & 64.7 \\
DFANet B               & 6  & \textbf{160}  & 59.3 \\
\hline
\end{tabular}
\end{center}
\caption{Results on CamVid \textit{test} set. }
\label{tab:CamVid Results}
\end{table}

\subsection{Comparison on Other Datasets}
We also evaluate our DFANet on CamVid dataset. 
CamVid contains images extracted from video sequences with resolution up to 960 $\times$ 720. 
It contains 701 images in total, in which 367 for training, 101 for validation and 233 for testing. We adopt the same setting as \cite{CamVid_set}.
The image resolution for training and evaluation are both 960 $\times$ 720. The results are reported in Table \ref{tab:CamVid Results}. DFANets get much faster inference speed 120 FPS and 160 FPS than other methods on this high resolution with slightly worse than the state-of-the-art methods\cite{ICNet}.

\section{Conclusion}
In this paper, we propose deep feature aggregation to tackle real-time semantic segmentation on high resolution image. Aggregation strategy connects a set of convolution layers to effectively refine high-level and low-level features, without any specifically designed operation. Analysis and quantitative experimental results on Cityscapes and CamVid dataset are presented to demonstrate the effectiveness of our method.

\textbf{Acknowledgements} This research was supported by National Key R\&D Program of China (No. 2017YFA0700800), and The National Key Research and Development Program of China (2018YFC0831700).

{\small
\bibliographystyle{ieee_fullname}
\bibliography{egbib}
}

\end{document}